%% file: Arxiv.tex
\documentclass[sigconf]{acmart}
\usepackage{threeparttable}
\usepackage{graphicx}
\usepackage{textcomp}
\usepackage{shortcuts}
\usepackage{framed}
\usepackage{xcolor}
\usepackage{url}
\usepackage{pifont}

\usepackage{breakurl}
\usepackage{enumitem}
\usepackage{multirow}

\usepackage[linesnumbered,titlenumbered,ruled,vlined,commentsnumbered]{algorithm2e}
\SetKwComment{Comment}{$\triangleright$\ }{}
\usepackage{comment}
\usepackage{enumitem}
\usepackage{dsfont}
\usepackage{booktabs}       
\usepackage{amsfonts}       
\usepackage{nicefrac}       
\usepackage{float}
\usepackage{microtype}      
\PassOptionsToPackage{citecolor=royalblue,breaklinks,colorlinks,pagebackref}{hyperref}

\usepackage{appendix}

\usepackage{newfloat}
\usepackage{listings}
\DeclareCaptionStyle{ruled}{labelfont=normalfont,labelsep=colon,strut=off} 
\floatstyle{ruled}
\newfloat{listing}{tb}{lst}{}
\floatname{listing}{Listing}

\definecolor{dg}{rgb}{0,0.694,0.298}
\definecolor{purple}{rgb}{0.4,0.176,0.569}
\definecolor{Gray}{gray}{0.6}
\definecolor{royalblue}{RGB}{65,105,225}

\makeatletter
\DeclareRobustCommand\onedot{\futurelet\@let@token\@onedot}
\def\@onedot{\ifx\@let@token.\else.\null\fi\xspace}
\def\eg{\emph{e.g}\onedot} 
\def\ie{\emph{i.e}\onedot}

\makeatother

\AtBeginDocument{%
  \providecommand\BibTeX{{%
    \normalfont B\kern-0.5em{\scshape i\kern-0.25em b}\kern-0.8em\TeX}}}

\setcopyright{acmcopyright}
\copyrightyear{2018}
\acmYear{2018}
\acmDOI{XXXXXXX.XXXXXXX}

\acmConference[Conference acronym 'XX]{Make sure to enter the correct
  conference title from your rights confirmation emai}{June 03--05,
  2018}{Woodstock, NY}
\acmPrice{15.00}
\acmISBN{978-1-4503-XXXX-X/18/06}
\acmSubmissionID{185}

\begin{document}

\title{Efficient and Effective Universal Adversarial Attack \\ against Vision-Language Pre-training Models}

\author{Fan Yang}
\affiliation{%
  \institution{Huazhong University of Science and Technology}
  \city{Wuhan}
  \country{China}
}

\author{Yihao Huang$^{\ast}$}
\affiliation{%
  \institution{Nanyang Technological University}
  \city{Singapore}
  \country{Singapore}
}
\thanks{*Corresponding authors. huangyihao22@gmail.com, wangkl@hust.edu.cn}

\author{Kailong Wang$^{\ast}$}
\affiliation{%
  \institution{Huazhong University of Science and Technology}
  \city{Wuhan}
  \country{China}
}

\author{Ling Shi}
\affiliation{%
  \institution{Nanyang Technological University}
  \city{Singapore}
  \country{Singapore}
}

\author{Geguang Pu}
\affiliation{%
  \institution{East China Normal University}
  \city{Shanghai}
  \country{China}
}

\author{Yang Liu}
\affiliation{%
  \institution{Nanyang Technological University}
  \city{Singapore}
  \country{Singapore}
}

\author{Haoyu Wang}
\affiliation{%
  \institution{Huazhong University of Science and Technology}
  \city{Wuhan}
  \country{China}
}

\begin{abstract}
Vision-language pre-training (VLP) models, trained on large-scale image-text pairs, have become widely used across a variety of downstream vision-and-language (V+L) tasks. This widespread adoption raises concerns about their vulnerability to adversarial attacks. Non-universal adversarial attacks, while effective, are often impractical for real-time online applications due to their high computational demands per data instance. Recently, universal adversarial perturbations (UAPs) have been introduced as a solution, but existing generator-based UAP methods are significantly time-consuming. To overcome the limitation, we propose a direct optimization-based UAP approach, termed DO-UAP, which significantly reduces resource consumption while maintaining high attack performance. Specifically, we explore the necessity of multimodal loss design and introduce a useful data augmentation strategy. Extensive experiments conducted on three benchmark VLP datasets, six popular VLP models, and three classical downstream tasks demonstrate the efficiency and effectiveness of DO-UAP. Specifically, our approach drastically decreases the time consumption by 23-fold while achieving a better attack performance.
\end{abstract}


\maketitle

\input{arxiv_doc}
\end{document}

%% file: arxiv_doc.tex
\noindent\textbf{Relevance.} We explore universal adversarial perturbations (UAPs) with a focus on their potential to reveal vulnerabilities in online vision-language applications. By leveraging UAPs' ability to deceive neural networks across various image-text pairs without the need for additional overhead, this study underscores critical robustness challenges in these online systems.

\section{Introduction}
In the evolving web ecosystem, where multimodal content is increasingly prevalent, vision-language pre-training (VLP) models~\cite{li2021align,yang2022vision,song2022clip,radford2021learning} play a crucial role in enhancing user experiences through applications like image search, automatic captioning, and content recommendation. They have transformed multimodal learning by utilizing large-scale image-text pairs to bridge visual and linguistic understanding. Models like CLIP \cite{radford2021learning}, ALBEF \cite{li2021align}, and TCL \cite{yang2022vision} have achieved remarkable results in vision-and-language (V+L) tasks \cite{wang2016comprehensive,Hong2019,Xie2019VisualEA,hu2022scaling}.
%
%
By employing self-supervised pre-training, these models align cross-modal features, enabling them to capture complex relationships between visual and textual data, thus generating effective multimodal representations.

Despite these advancements, recent research  \cite{zhao2024evaluating,gao2024boosting,liu2024multimodal,zhang2022towards,lu2023set} have identified \textbf{critical vulnerabilities in VLP models}, particularly when faced with multimodal adversarial examples. 
%
%
However, these approaches typically rely on generating specific adversarial perturbations for each image-text instance, requiring new perturbations to be learned from scratch. This leads to \textbf{significant computational overhead}, limiting the applicability of such attacks in practical scenarios like real-time online applications. 

\begin{figure}[tb]
\centering
        \includegraphics[width=\linewidth]{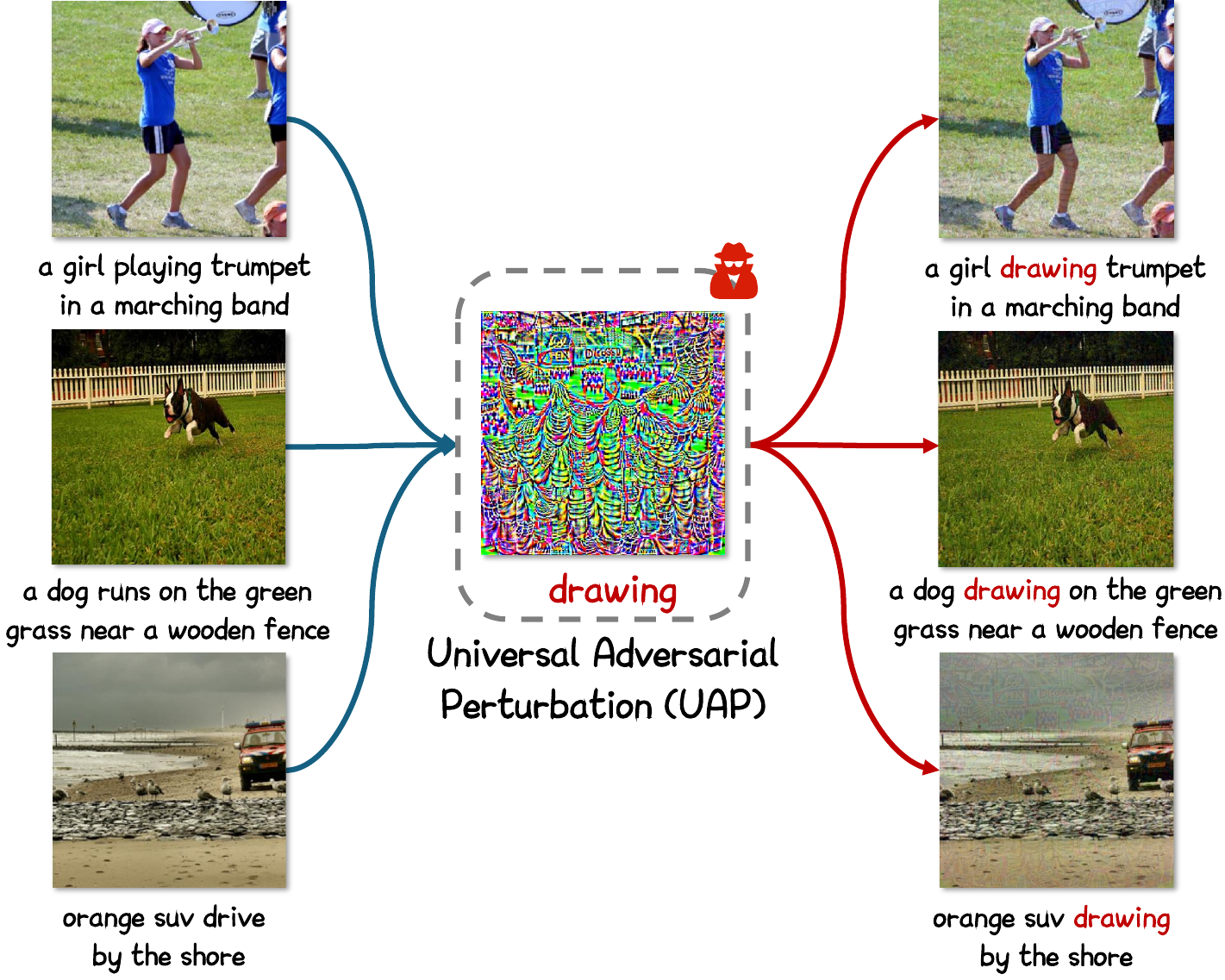}
	\caption{Effect of the universal adversarial perturbations against VLP models. With just a pair of \textit{fixed} image-text perturbations, the proposed attack effectively misleads arbitrary image-text pairs across a wide range of V+L tasks.}
	\label{fig:teaser}
\vspace{-20pt}
\end{figure}

To make the multimodal attack perturbation more practical, designing the universal adversarial perturbation (UAP) method (Fig.~\ref{fig:teaser}) is a promising direction. There are two basic UAP generation pipelines: \textit{direct optimization-based} \cite{moosavi2017universal} and \textit{generator-based} \cite{poursaeed2018generative}.
Through an empirical study, the recent work CPGC \cite{fang2024one} finds the \textit{direct optimization-based} pipeline simply updates UAPs via gradients but often lacks effectiveness. Comparatively, the \textit{generator-based} pipeline is more effective for universally attacking VLP models. As a result, they introduce the first generator-based multimodal UAP designed to mislead VLP models using \textbf{fixed, data-agnostic} perturbations. 
Despite achieving high performance (\textasciitilde 90\%), the generator-based approach of CPGC is inefficient, demanding significant time and data resources (roughly one week on a single A100 GPU). This inefficiency motivates us to investigate a crucial question: ``Can we design a more efficient, direct optimization-based UAP method to achieve comparable effectiveness against VLP models?''

To achieve this, an intuitive idea is to utilize a cross-modal image-text loss to generate UAPs that disrupt the image-text alignment. However, upon further analysis, we reveal that this approach is insufficient due to the ``modality gap'' phenomenon observed in VLP models \cite{liang2022mind}. The issue arises because the perturbed image-text pairs remain close to their unperturbed counterparts, which limits their universal effectiveness. To address this limitation, we highlight the necessity of incorporating an unimodal loss alongside the basic cross-modal image-text loss. 
Additionally, while data augmentations are commonly used to improve adversarial attack performance \cite{lu2023set, xie2019improving}, their impact on UAP attacks remains unclear, and not all augmentations are beneficial for UAP generation. To this end, we introduce an image-text similarity-based idea for selecting useful data augmentations (\ie, brightness). Our approach drastically decreases the time consumption by 23-fold compared to CPGC~(state-of-the-art). 

To sum up, our work has the following contributions:
\begin{itemize}[itemsep=2pt,topsep=0pt,parsep=0pt]
\item To the best of our knowledge, we propose the first \textit{direct optimization-based} multimodal UAP method targeting VLP models. The approach significantly outperforms generator-based methods, delivering superior attack performance while reducing time consumption.
\item We propose a multimodal-oriented loss function and incorporate carefully selected data augmentation, both of which significantly enhance the attack performance of UAP.
\item The experimental results on three benchmark VLP datasets, six widely-used VLP models, and three classical downstream tasks strongly affirm the effectiveness of DO-UAP, underscoring the robustness and adaptability of our method across diverse settings.
\end{itemize}

\section{Background and Related Work}
\subsection{Vision-Language Pre-training Models}
Vision-language pre-training (VLP) is a key technique for improving multimodal tasks by utilizing large-scale image-text pairs \cite{chen2023vlp}. Traditionally, research relied on pre-trained object detectors for multimodal representation \cite{li2021align,yang2022vision}, but Vision Transformers (ViT) \cite{dosovitskiy2020image} have shifted the approach toward end-to-end image encoding, transforming inputs into patches \cite{han2022survey}.

VLP models are divided into two main types: fused and aligned. Fused models, like ALBEF \cite{li2021align} and TCL \cite{yang2022vision}, use separate unimodal encoders for text and images, followed by a multimodal encoder to combine the embeddings. In contrast, aligned models, such as CLIP \cite{radford2021learning}, use unimodal encoders to independently process image and text embeddings, aligning them for downstream tasks.

For instance, CLIP leverages contrastive learning to align embeddings by maximizing the similarity of matched image-text pairs while minimizing that of unmatched ones. Models like BLIP \cite{li2022blip} improve training by synthesizing and filtering captions for web images. Fused models like ALBEF align and then merge unimodal representations using cross-modal attention, while TCL employs contrastive learning to maintain global-local alignment. These pre-trained models show great generalizability and can be fine-tuned for various applications.

\subsection{Downstream Vision-and-Language Tasks}
Image-Text Retrieval (ITR) involves retrieving relevant information from a database using one modality (image or text) to query the other. It includes image-to-text retrieval (TR) and text-to-image retrieval (IR) \cite{ijcai2022p759,zhang2020context}. Fused models like ALBEF and TCL rank results based on multimodal encoders, while models such as CLIP rank using unimodal embedding similarity \cite{radford2021learning}. Fused models retrieve Top-N candidates for final ranking via a multimodal encoder.

Image Captioning (IC) \cite{bai2018survey,vinyals2015show} generates textual descriptions of visual content, translating images into coherent captions. Unlike ITR, IC focuses on describing visual input, often evaluated using metrics like BLEU and CIDEr.

Visual Grounding (VG) \cite{Hong2019,li2020visual,shrestha2020negative} focuses on localizing objects or regions in an image based on text descriptions.


\subsection{Universal Adversarial Perturbation}
Research on the safety and security of VLP models \cite{li2024safegen,wu2024legilimens,deng2025raconteur,liu2024compromising,zhou2024foolsdedit,zhou2024mip} has gained significant attention, with prior studies \cite{kim2019single,shah2019cycle,xu2018fooling} predominantly focusing on unimodality perturbations, thus limiting focus on cross-modal interactions.

To overcome this, Co-Attack \cite{zhang2022towards} introduces a multimodal adversarial attack for VLP models, targeting image-text interactions to disrupt embedding distances between adversarial examples and original data pairs. It is applicable to models like CLIP, ALBEF, and TCL. SGA \cite{lu2023set} enhances transferability by using set-level augmentations and cross-modal guidance, increasing data diversity and reducing similarity between adversarial examples and their matched modality. However, it remains instance-specific and limited in real-time applications. TMM \cite{wang2024transferable} improves on SGA by better leveraging cross-modal interactions through optimizing modality-consistency and modality-discrepancy features.

To improve the applicability and generalizability in practice, several works have explored UAP for VLP models. Zhang et al. \cite{zhang2024universal} proposed a pseudo-universal attack by applying UAP solely to the image modality, which does not constitute a truly universal attack for multimodal VLP models. In contrast, CPGC \cite{fang2024one} introduced the first fully universal attack for VLP models by adding perturbations to both image and text modalities. Through contrastive training, the generator disrupts the alignment between image-text pairs, enhancing attack effectiveness across various vision-language tasks and models. However, their method involves optimizing a generator for UAP generation, which is time-consuming and unstable. To address this, we propose a more efficient UAP generation method using direct optimization.

\section{Methodology}
In this section, we begin by outlining the problem of universal adversarial attacks on VLP models, followed by a detailed introduction to our direct optimization method for UAP generation.

\subsection{Problem Statement}
\subsubsection{Problem Definition}
Let ($v$,$t$) represent an image-text pair from a multimodal dataset $\mathcal{D}$. The image encoder $f_{I}(\cdot)$ and text encoder $f_{T}(\cdot)$ are components of a target VLP model $f(\cdot)$. The objective of multimodal universal adversarial attacks is to generate \textbf{fixed} image-text perturbations ($\delta_v,\delta_t$) that can be applied universally across a large subset of ($v$,$t$) pairs from the test dataset $\mathcal{D}_{test}$, misleading the VLP model $f(\cdot)$ into making incorrect decisions.

\subsubsection{Preliminary} The multimodal perturbations ($\delta_v,\delta_t$) are typically constrained to small magnitudes ($\epsilon_v,\epsilon_t$). For the image modality, we follow the settings of prior work \cite{wang2024transferable,fang2024one} and adopt the strength $\epsilon_v$ of $\delta_v$ to be the value of 12/255 using the $l_{\infty}$ norm. For the text modality, we adopt the setting of the previous work \cite{zhang2022towards,lu2023set,wang2024transferable,fang2024one}, where key tokens in the original text $t$ are replaced with crafted adversarial tokens. The textual perturbation is thus token-level, with $\epsilon_t$ restricted to substituting only one token per sentence (\ie, $\epsilon_t = 1$). 

\begin{figure}[tb]
\centering
        \includegraphics[width=\linewidth]{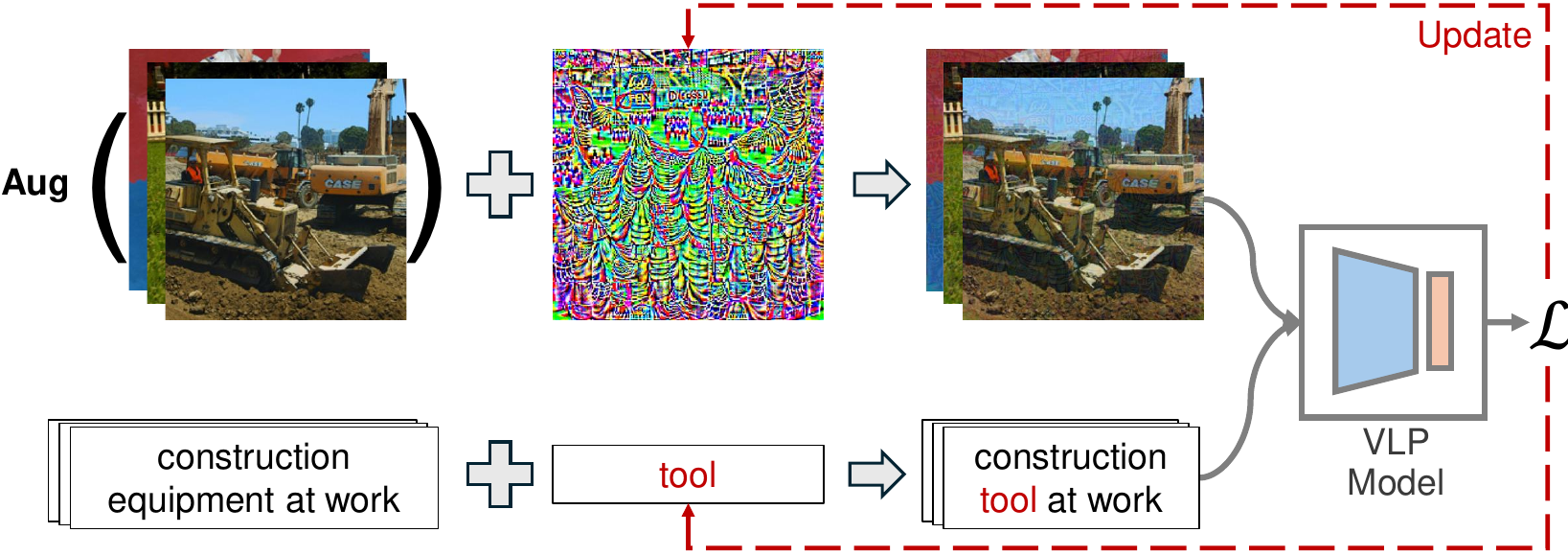}
	\caption{Pipeline of DO-UAP method.}
	\label{fig:pipeline}
\end{figure}

\subsection{Method Design Intuition}
\begin{figure}[tbp]
\centering
        \includegraphics[width=0.7\linewidth]{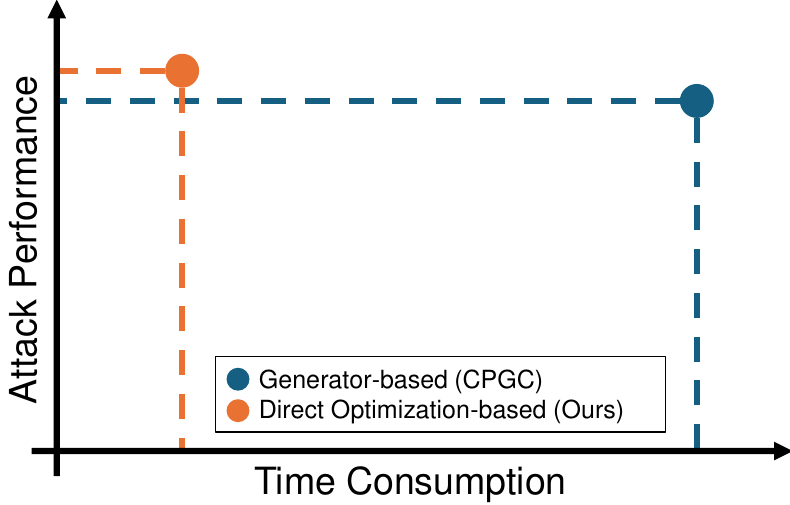}
	\caption{Compared to existing generator-based UAP methods, our proposed direct optimization-based UAP approach not only delivers higher attack performance but also significantly reduces time consumption.}
	\label{fig:coarse_compare_baseline}
\end{figure}

Based on the empirical finding that generator-based attacks \cite{poursaeed2018generative} surpass direct optimization-based attacks \cite{moosavi2017universal} in compromising VLP models, CPGC \cite{fang2024one} proposes a generator-based UAP method specifically targeting VLP models. While this choice makes them easily achieve high attack performance, \textit{they overlook a key drawback of generative methods: optimizing a generator for UAP generation is an indirect, time-consuming process.} In contrast, directly optimizing the UAP with respect to the loss function is a more efficient and resource-saving solution (see Figure~\ref{fig:coarse_compare_baseline}).

To develop a direct optimization-based method for universal adversarial attacks, an intuitive idea is to adopt the objective design utilized in previous universal adversarial attack methods \cite{poursaeed2018generative, shafahi2020universal, TSC-UAP} aimed at unimodality tasks, \ie, maximizing the task-specific loss. Note that for clarity and conciseness, we use the image classification task as a representative example of an unimodality task. Specifically, in traditional image classification, the adversarial objective typically involves maximizing the cross-entropy loss between the model's prediction label and the ground truth label.

Similarly, for universal adversarial attacks targeting VLP models, as the V+L tasks (\eg, image-text retrieval, image captioning, visual grounding) are all cross-modal tasks that rely on the alignment between image and text, the basic objective is to maximize the loss $\mathcal{L}_t = \mathcal{J}(f_{I}(v'),f_{T}(t'))$ to disrupts the image-text alignment: 
\begin{align}
\max ~ \mathcal{L}_t, ~ \textit{s.t.} ~ v' \in B[v,\epsilon_v], ~ t' \in B[t,\epsilon_t],
\label{con:goal}
\end{align}
where $v' = v + \delta_v$ representing the adversarial image and $t' = t \oplus \delta_t$ denoting adversarial text ($\oplus$ means replace token). $\mathcal{J}(\cdot)$ is the training loss used in the VLP models. The allowable perturbations of ($v$,$t$) are restricted within the ranges $B[v,\epsilon_v]$ for images and $B[t,\epsilon_t]$ for text.

\textbf{Method Overview.} Following Eq.\eqref{con:goal}, the generated UAP demonstrates some level of attack effectiveness. However, the ``modality gap'' phenomenon in VLP is not fully considered, limiting the comprehensiveness of the attack. To this end, we propose a multimodal-oriented loss design in Sec.~\ref{sec:moe}. Furthermore, to further boost the attack performance of our UAP method, we propose an idea for selecting a suitable data augmentation method, identifying brightness augmentation as a particularly effective choice, as detailed in Sec.~\ref{sec:das}. The pipeline of our DO-UAP method is in Figure~\ref{fig:pipeline}. Given the image-text pairs, we first perform data augmentation on a batch of images for the image modality. Adversarial images are generated by adding UAP to these augmented images. For the text modality, key tokens in the original text are replaced with adversarial tokens. Both adversarial images and texts are then fed into the VLP model to compute the loss. This loss is subsequently used to update the UAP for both image and text modalities.

\subsection{Multimodal-Oriented Loss Design}\label{sec:moe}
Given that the purpose of VLP models is to align image-text pairs, it is logical to design a cross-modal objective for universal adversarial attacks. However, the Eq.~\eqref{con:goal} is inadequate when accounting for the complexities of multimodal models.
The complexity arises from the ``modality gap'' phenomenon in VLP models \cite{liang2022mind}, where the embedding vectors of images and text are distinctly separated within the joint vision-language embedding space, as observed in VLP models like CLIP. If there is no ``modality gap'' phenomenon, the cross-modal objective is good enough.

\begin{figure}[tb]
\centering
        \includegraphics[width=0.7\linewidth]{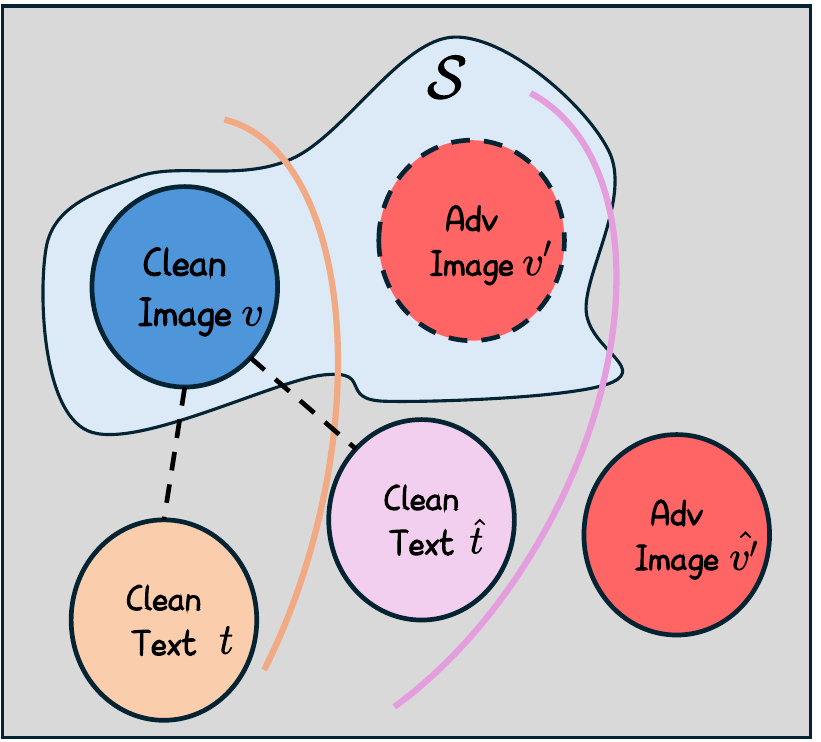}
	\caption{Motivation for multimodal loss design.}
	\label{fig:single_modality_loss_motivation}
\end{figure}

To be specific, we give an explanation in Figure~\ref{fig:single_modality_loss_motivation} to demonstrate the problem in Eq.~\eqref{con:goal} raised by this phenomenon. For simplicity and clarity, we focus solely on the UAP for image (\ie, $\delta_v$) though a similar reasoning applies to the text modality. In Figure~\ref{fig:single_modality_loss_motivation}, the image-text pair ($v$,$t$) from the original training dataset are in blue and orange circles respectively. In the image domain, the space $\mathcal{S}$ represents the set of images with semantics similar to image $v$. The orange line indicates the semantic boundary, determining whether the semantics of an image align closely with the clean text $t$. When optimizing for a cross-modal objective, the semantics of the generated adversarial image $v'$ (\ie, $v + \delta_v$) diverge significantly from the text $t$ (on both sides of the orange boundary), but may still reside within the space $\mathcal{S}$ (represented by the red circle with a dotted outline). 
%
As the adversarial image $v'$ retains similar semantics to the original image $v$, in comparison to other texts in the test dataset (e.g., $\hat{t}$, represented by the purple circle) which correspond to the clean image $v$ but differ semantically from $t$ (purple \textit{vs.} orange boundary), the VLP model is likely to retain the alignment of the adversarial image $v'$ with $\hat{t}$. 
This results in poor attack performance of UAP $\delta_v$ generated solely based on the cross-modal objective. 

To address the problem and enhance the attack performance of the UAP $\delta_v$, we propose to add an unimodal objective. As shown in Figure~\ref{fig:single_modality_loss_motivation}, adding an unimodal objective into the clean image can push the adversarial image further away from the space $\mathcal{S}$, transforming it to be $\hat{v'}$ (see a red circle with solid outline). Since adversarial image $\hat{v'}$ is positioned outside space $\mathcal{S}$,  there is a high probability that the image $\hat{v'}$ and text $\hat{t}$ are not aligned. 

To summarize, we propose the necessary of a multimodal objective combining cross-modal loss and unimodal loss, expanding the basic objective Eq.~\eqref{con:goal} to:
\begin{align}
\mathcal{L}_u &= \mathcal{D}(f_{I}(v'), f_{I}(v)) + \mathcal{D}(f_{I}(t'), f_{I}(t))\\
\mathcal{L} &= \mathcal{L}_t + \alpha \mathcal{L}_u \label{al:L}\\
\max & ~ \mathcal{L}, ~ \textit{s.t.} ~ v' \in B[v,\epsilon_v], ~ t' \in B[t,\epsilon_t],
\label{con:goal_multi_modal}
\end{align}
where function $\mathcal{D}(\cdot)$ is used to evaluate the distance between embeddings and we use cosine similarity here. $\alpha$ is the weight.

\subsection{Data Augmentation Selection}\label{sec:das}
To further enhance the performance of UAP attacks, we leverage data augmentation, a well-established technique in deep learning \cite{zhong2020random, rebuffi2021data, wang2024comprehensive}. The impact of data augmentation on UAP attacks specifically targeting VLP models has not yet been explored. Given that not all types of data augmentation are beneficial \cite{cubuk2021tradeoffs, shen2022data}, we aim to find an effective data augmentation for UAP generation. 

Our high-level idea stems from analyzing the attack objective on VLP models (\ie, Eq.~\eqref{con:goal}), where the goal of adversarial perturbation is designed to disrupt the alignment between text and image pairs.
Let's consider a special case where \textbf{misaligned image-text pairs} (\ie, with low image-text similarity, such as an image of a dog paired with a text description of a computer) are used as input, instead of the typically aligned pairs (\ie, with high image-text similarity, such as an image of a dog paired with a text description of a dog).
Intuitively, when misaligned image-text pairs are used for UAP generation, their inherent misalignment already satisfies the attack objective and does not need UAP. As a result, the gradients calculated by such misaligned image-text pairs become less informative and meaningless for the UAP generation. Building on this insight, we suppose that data augmentations that enhance the similarity between image-text pairs can positively impact UAP generation. To this end, we conduct an empirical study on five data augmentation types (\ie, compression, noise, flip, crop, and brightness) and try to find one that is effective for UAP generation. 
Note that in this study, we select these five types of data augmentation due to their widespread use in a variety of tasks. However, this choice does not imply that other augmentation techniques are unsuitable for generating UAPs. Additionally, our objective is not to exhaustively identify all effective data augmentation methods for UAP generation.

%
To evaluate the influence of data augmentations on image-text similarity, we apply them to the test dataset of Flickr30K \cite{plummer2015flickr30k} targeting CLIP$_{\text{ViT}}$ model. The results are shown in Table~\ref{tab:augmentation_similarity}. In the first row are the names of data augmentations and ``origin'' means no augmentation. In the ``image-text similarity'' row, we give the average value of image-text similarity in the test dataset of Flickr30K by calculating the cosine similarity of the image-text pair ($v$,$t$), \ie, $CosineSim(f_{I}(v),f_{T}(t))$. From the table, we can find that only the ``brightness'' augmentation can improve the image-text similarity while other augmentation methods (\eg, ``flip'') are not effective, or even reduce the similarity (\eg, ``compression''). 

We further generate the UAP by attacking the CLIP$_{\text{ViT}}$ model on the Flickr30K dataset with these data augmentation methods based on Eq.~\eqref{al:L} and evaluate the performance of the generated UAP on the image-to-text retrieval (ITR) task. Note that for the ITR task, there are actually two tasks, image retrieval (IR) and text retrieval (TR). We report the average attack performance across both tasks in Table~\ref{tab:augmentation_similarity}. The evaluation metric employed is the attack success rate (ASR) on Recall@1 (R@1). R@1 is a standard metric in Image-Text Retrieval (ITR) tasks, where a score of 1 is assigned if the system retrieves a relevant result in the first position for a given query, and 0 otherwise. For UAP, a higher ASR indicates better performance. In the ``attack performance'' row of Table~\ref{tab:augmentation_similarity}, we observe that only the UAP generated with ``brightness'' augmentation outperforms the version without augmentation. Augmentations that reduce image-text similarity result in worse attack performance of their corresponding UAP. 

In summary, from the empirical study, we suggest adding brightness augmentation to the UAP generation algorithm. We also observe a positive correlation between the effect of data augmentation on image-text similarity and its impact on UAP performance. 
It is important to note that this phenomenon is observed with only certain types of data augmentation and may not be effective across all augmentation methods. As our primary objective is to identify a suitable augmentation for UAP generation, we did not pursue further exploration in this direction. However, we believe this is an interesting avenue for future research. 
%

\begin{table}[tbp]
\centering
\caption{Evaluating the impact of various data augmentations on image-text similarity and the attack performance of UAP.}
\resizebox{\linewidth}{!}{
\begin{tabular}{l|c|cccccc}
\toprule 
 & Origin & Compression & Noise & Flip & Crop & Brightness\tabularnewline
\midrule 
Image-text similarity & 0.4562 & 0.4554 & 0.4554 & 0.4562 & 0.4484 & \textbf{0.4608}\tabularnewline
Attack performance $\uparrow$ & 92.765 & 89.155 & 88.125 & 92.640 & 90.210 & \textbf{94.360}\tabularnewline
\bottomrule 
\end{tabular}
}
\label{tab:augmentation_similarity}
\end{table}
\begin{algorithm}[tb]
	{
		\caption{DO-UAP Algorithm}\label{alg:alg_our_UAP}
		\KwIn{Training dataset $\mathcal{D}$, Loss $\mathcal{L}$, Batch size $m$, Number of epochs $E$, Perturbation magnitude $\epsilon_v$ and $\epsilon_t$}
		\KwOut{Perturbation $\delta_v$ and $\delta_t$}
            $\delta_v, \delta_t \gets 0$ \label{line:UAP_init} \Comment*[r]{initialization}
            \For{$epoch = 1\ \mathrm{to}\ {E}$}
            {\label{line:epoch}
                $I \gets |\mathcal{D}|/m$ \label{line:cal_iteration} \Comment*[r]{iteration number}
     		\For{$iteration = 1\ \mathrm{to}\ {I}$}{\label{line:iteration}
     		    $\mathcal{P} \sim \mathcal{D}: |\mathcal{P}|=m$ \label{line:sample_batch} \Comment*[r]{randomly sample}
                    $\mathcal{P} \gets Aug(\mathcal{P})$ \label{line:aug} \Comment*[r]{brightness augmentation}
     		    $g_{(\delta_v,\delta_t)} \gets \mathop{\mathds{E}}\limits_{(v,t) \sim \mathcal{P}}[\nabla_{(\delta_v,\delta_t)}\mathcal{L}]$ \label{line:cal_gradient} \Comment*[r]{$\mathcal{L}$ from Eq.~\eqref{al:L}}
     		$\delta_v,\delta_t \gets \mathrm{Optim}(g_{(\delta_v,\delta_t)}$)\label{line:update_UAP} \Comment*[r]{update UAP}
                $ \mathrm{let}~v+\delta_v \in B[v, \epsilon_v], ~ t \oplus \delta_t \in B[t, \epsilon_t]$ \label{line:clip_UAP} \Comment*[r]{clipping}
                }
            }
	}
\end{algorithm}
\begin{table*}[tbp]
\centering
\caption{Compare with baselines.}
\resizebox{\linewidth}{!}{
\begin{tabular}{c|c|cccccccccccc|cc|c}
\toprule 
\multirow{2}{*}{Dataset} & \multirow{2}{*}{Methods} & \multicolumn{2}{c}{CLIP$_{\text{ViT}}$} & \multicolumn{2}{c}{CLIP$_{\text{CNN}}$} & \multicolumn{2}{c}{BLIP} & \multicolumn{2}{c}{TCL} & \multicolumn{2}{c}{X-VLM} & \multicolumn{2}{c|}{ALBEF} & \multicolumn{2}{c|}{Average} & \multirow{2}{*}{Time (hours)}\tabularnewline
\cline{3-16}
 &  & TR & IR & TR & IR & TR & IR & TR & IR & TR & IR & TR & IR & TR & IR & \tabularnewline
\midrule 
\multirow{4}{*}{Flickr30K} & GAP & 81.63 & 83.91 & 76.11 & 85.20 & 48.16 & 73.70 & 84.20 & 80.34 & 80.35 & 76.91 & 72.46 & 82.69 & 73.82 & 80.46 & 5.22\tabularnewline
 & CPGC-2 & 91.13 & 93.75 & 80.18 & 87.15 & 69.93 & 78.49 & 91.72 & 87.46 & 81.81 & 82.19 & 87.67 & 84.92 & 83.74 & 85.66 & 6.71\tabularnewline
 & CPGC-40 & 89.04 & 93.59 & 82.12 & 88.94 & 80.63 & 87.31 & \textbf{95.45} & \textbf{91.57} & 93.70 & 92.20 & 90.24 & 89.51 & 88.53 & 90.52 & 134.19\tabularnewline
 & DO-UAP (ours) & \textbf{93.35} & \textbf{95.37} & \textbf{93.13} & \textbf{93.84} & \textbf{88.33} & \textbf{91.98} & 94.10 & 91.36 & \textbf{97.36} & \textbf{94.55} & \textbf{94.45} & \textbf{93.55} & \textbf{93.45} & \textbf{93.44} & 5.83\tabularnewline
\midrule 
\multirow{4}{*}{MSCOCO} & GAP & 97.79 & 96.38 & 91.43 & 94.47 & 75.17 & 73.76 & 95.66 & 92.98 & 95.47 & 89.53 & 83.63 & 84.82 & 89.86 & 88.66 & 5.21\tabularnewline
 & CPGC-2 & 96.07 & 96.91 & 95.06 & 95.09 & 88.14 & 86.36 & 94.31 & 87.60 & 96.65 & 92.01 & 85.87 & 83.58 & 92.68 & 90.26 & 6.51\tabularnewline
 & CPGC-40 & \textbf{98.97} & \textbf{98.14} & 95.30 & 94.59 & \textbf{95.84} & \textbf{95.86} & 97.20 & 94.35 & \textbf{99.04} & \textbf{96.07} & 96.49 & 95.47 & \textbf{97.14} & \textbf{95.75} & 130.13\tabularnewline
 & DO-UAP (ours) & 98.13 & 98.01 & \textbf{96.90} & \textbf{98.01} & 88.38 & 87.31 & \textbf{98.73} & \textbf{96.51} & 98.23 & 96.02 & \textbf{96.73} & \textbf{97.23} & 96.18 & 95.52 & 5.63\tabularnewline
\bottomrule 
\end{tabular}
}
\label{tab:baseline_comparasion}
\end{table*}

\subsection{Implementation}

Our proposed approach DO-UAP is implemented following 
Algorithm~\ref{alg:alg_our_UAP}. In line~\ref{line:UAP_init}, the UAP pair $\delta_v, \delta_t$ are initialized. In line~\ref{line:epoch}, executes the adversarial attack over $E$ epochs. In line~\ref{line:cal_iteration}, the number of iterations $I$ is determined. In line~\ref{line:iteration}, carries out the adversarial attack iteratively for $I$ iterations. In line~\ref{line:sample_batch}, a random sample of training data $\mathcal{P}$ is selected with a fixed batch size $m$. In line~\ref{line:aug}, apply brightness augmentation to data. In line~\ref{line:cal_gradient}, the gradient $g_{(\delta_v,\delta_t)}$ for the UAP is calculated. In line~\ref{line:update_UAP}, update the UAP pair ($\delta_v, \delta_t$). In line~\ref{line:clip_UAP}, clipping the UAP ($\delta_v, \delta_t$) to satisfy the constraint.

\section{Experiment}
\subsection{Experimental Setup}
\paragraph{Downstream tasks and datasets}
We conducted a comprehensive study of our method across three downstream V+L tasks, namely image-text retrieval (ITR), image captioning (IC), and visual grounding (VG). The datasets used for these tasks are as follows:
\begin{itemize}[leftmargin=10pt, itemsep=0pt,topsep=0pt,parsep=0pt]
\item The Flickr30K dataset \cite{plummer2015flickr30k}, sourced from the Flickr website, provides detailed descriptions of objects and activities, making it a widely used benchmark for various V+L tasks. It comprises 31,783 images, each accompanied by five captions. We utilize this dataset for ITR tasks. 
\item The MSCOCO dataset \cite{lin2014microsoft} is a comprehensive and diverse collection containing 123,287 images, each annotated with approximately five descriptive sentences. This dataset is employed to assess attack performance in both ITR and IC tasks.
\item The RefCOCO+ dataset \cite{yu2016modeling}, a subset of MSCOCO, includes 19,992 images and 141,564 annotations, specifically designed for VG tasks.
\end{itemize}
\paragraph{Surrogate models and victim models} For the ITR tasks, we employ CLIP$_{\text{ViT}}$, CLIP$_{\text{CNN}}$ \cite{radford2021learning}, BLIP \cite{li2022blip} ALBEF \cite{li2021align}, TCL \cite{yang2022vision}, and X-VLM \cite{zeng22c}, as victim models. Notably, different image feature extraction modules are used for CLIP: ViT-B/16 for CLIP$_{\text{ViT}}$ and ResNet-101 for CLIP$_{\text{CNN}}$, while the other models rely solely on ViT for image feature extraction.
%
For the VG tasks, ALBEF, TCL, and X-VLM serve as both surrogate and victim models. For the IC tasks, BLIP is used as the victim model. Notably, out of the six VLP models, only ALBEF, TCL, and X-VLM can handle VG tasks, while only BLIP is capable of managing IC tasks.

\paragraph{Metrics} We use the attack success rate (ASR) as the primary metric to assess the effectiveness of our universal adversarial samples. ASR measures the extent to which adversarial perturbations cause deviations from correct decisions in the victim models. In ITR tasks, metrics such as R@1, R@5, and R@10 are typically used to calculate ASR, where R@N denotes the top N most relevant text or image matches based on a query. Due to space constraints, we only report the ASR for R@1 in our main results. For other cross-modal downstream V+L tasks, including IC and VG, their respective evaluation metrics will be outlined when presenting the experimental results in subsequent sections.

\paragraph{Baselines} We adopt the widely recognized and powerful GAP algorithm \cite{poursaeed2018generative} to the multimodal attack scenario by modifying its original loss function. Additionally, we compare our method with state-of-the-art (SOTA) methods CPGC \cite{fang2024one}. 

\paragraph{Implementation details} In line with prior works \cite{fang2024one}, we utilize the Karpathy split \cite{papineni2002bleu} to preprocess the dataset and construct the test set for performance evaluation. Also, following \cite{fang2024one}, for Flickr30K and MSCOCO, we randomly select 30,000 images along with their corresponding captions from the training set. For SNLI-VE and RefCOCO+, we directly utilize their respective training sets, consisting of 29,783 and 16,992 images. Regarding the perturbation budget, we follow the setting of previous work \cite{wang2024transferable,fang2024one}, constraining the visual perturbation $\epsilon_v$ to 12/255 and limiting the language perturbation $\epsilon_t$ to 1 to maintain invisibility. We set the number of training epochs for our DO-UAP to 2, using the brightness data augmentation range of [0, 0.05]. For the parameter $\alpha$ in Eq.~\eqref{al:L}, which balances the cross-modal and unimodal losses, we set $\alpha = 1$. For the attack strength per iteration, we set it to 0.1 of the perturbation budget. All the experiments are run on a Ubuntu system with an NVIDIA A100 Tensor of 80G RAM.

\subsection{Compare with SOTA Baseline}\label{sec:SOTA_comparison}
In alignment with previous adversarial attacks on VLP models \cite{zhang2022towards,lu2023set,fang2024one}, our focus is on the multimodal task of image-text retrieval, rather than image captioning or visual grounding tasks. This encompasses both image retrieval (IR) and text retrieval (TR). We conduct extensive experiments on the Flickr30K and MSCOCO datasets to assess the attack efficacy of the proposed DO-UAP. The experimental results across six different VLP models are summarized in Table~\ref{tab:baseline_comparasion}. We report the attack success rate (higher is better) on R@1 (results on R@5 and R@10 are in Appendix~\ref{sec:more_results_ITR}). We introduce a comparison with the basic generator-based UAP methods GAP \cite{poursaeed2018generative} and the state-of-the-art UAP method CPGC \cite{fang2024one}. In Table~\ref{tab:baseline_comparasion}, it is important to note that while CPGC optimizes its generator over 40 epochs (``CPGC-40'' row) in the original work, our DO-UAP method requires only 2 epochs, resulting in considerable time savings. For a fair comparison, we also limit CPGC to 2 epochs, with the results presented in the ``CPGC-2'' row. We also tracked the time consumption and present the average values here, with detailed records provided in Section~\ref{sec:time_consumption}.

Regarding the Flickr30K dataset, our method surpasses CPGC-40, achieving an average improvement of 4.92\% on the TR task and 2.92\% on the IR task across six models. Additionally, it significantly outperforms CPGC-2 and GAP by nearly 10\%. Notably, our DO-UAP method completes UAP generation in just 5.83 hours, significantly faster than the time-consuming CPGC-40, which takes an average of 134.19 hours across the same six models, representing a time efficiency improvement of approximately 23-fold.

Regarding the MSCOCO dataset, our method performs comparably to CPGC-40, with a slight average decrease of 0.96\% in the TR task and 0.23\% in the IR task across six models. However, it significantly surpasses CPGC-2 and GAP by nearly 5\%. Notably, our method generates UAP in just 5.63 hours, significantly faster than the 130.13 hours required by CPGC-40 across the same six models, representing a roughly 23-fold improvement in time efficiency. 

In summary, across both Flickr30K and MSCOCO datasets, our DO-UAP method surpasses CPGC in attack performance, with an improvement of 1.98\% on the TR task and 1.33\% on the IR task. More importantly, our DO-UAP method is stably more efficient than CPGC, being 20 times faster. Specifically, our method generates UAP in just six hours, while CPGC requires over five days.

\subsection{Time Consumption}\label{sec:time_consumption}
\begin{table}[tbp]
\centering
\caption{Comparison of time consumption.}
\resizebox{\linewidth}{!}{
\begin{tabular}{c|c|cccccc}
\toprule
\multirow{1}{*}{Dataset} & Methods & CLIP$_{\text{ViT}}$ & CLIP$_{\text{CNN}}$ & BLIP & TCL & X-VLM & ALBEF\tabularnewline
\midrule 
\multirow{4}{*}{\rotatebox{90}{Flickr30K}} & GAP & 2.86  & 2.14  & 6.50  & 6.14  & 7.78  & 5.92 \tabularnewline
 & CPGC-2 & 4.06  & 3.54  & 9.56  & 7.58  & 7.84  & 7.68 \tabularnewline
 & CPGC-40 & 81.20  & 70.80  & 191.20  & 151.60  & 156.80  & 153.52 \tabularnewline
 & DO-UAP (ours) & 3.80  & 2.94  & 6.94  & 7.22  & 7.24  & 6.84 \tabularnewline
\midrule
\multirow{4}{*}{\rotatebox{90}{MSCOCO}} & GAP & 2.84  & 2.12  & 6.20  & 6.52  & 7.80  & 5.78 \tabularnewline
 & CPGC-2 & 3.82  & 3.52  & 8.94  & 7.42  & 7.84  & 7.50 \tabularnewline
 & CPGC-40 & 76.40  & 70.40  & 178.80  & 148.40  & 156.80  & 150.00 \tabularnewline
 & DO-UAP (ours) & 3.60  & 2.66  & 6.80  & 6.92  & 7.10  & 6.68 \tabularnewline
\bottomrule 
\end{tabular}
}
\label{tab:time_comparasion}
\end{table}
\begin{figure*}[tb]
\centering
        \includegraphics[width=0.9\linewidth]{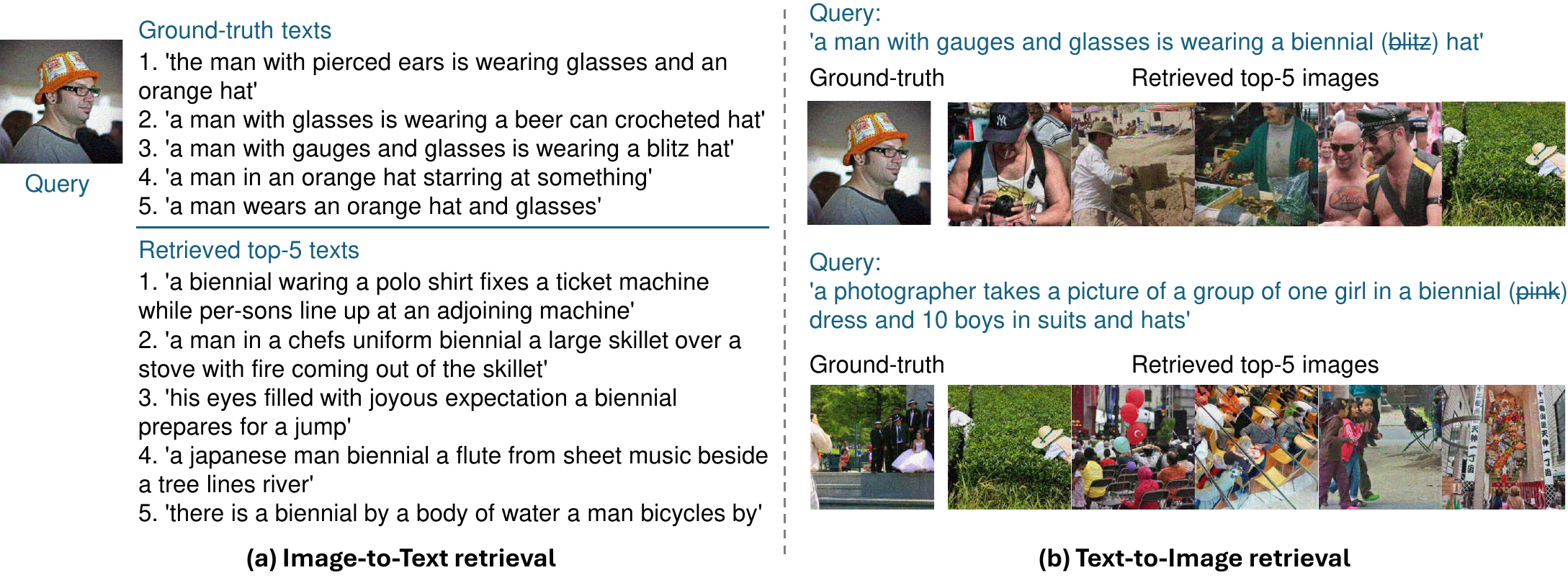}
	\caption{Visualization of image-text retrieval task.}
	\label{fig:ITR_visualization}
\vspace{-10pt}
\end{figure*}
\begin{figure}[tb]
\centering
        \includegraphics[width=\linewidth]{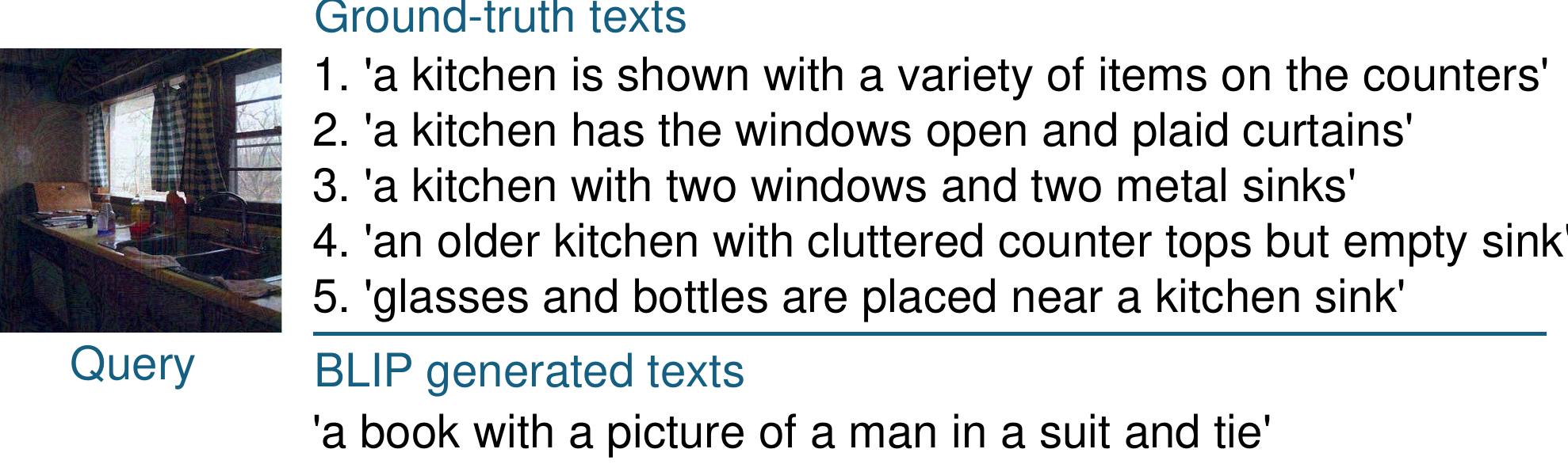}
	\caption{Visualization of image caption task.}
	\label{fig:image_caption}
\vspace{-10pt}
\end{figure}
\begin{figure}[tb]
\centering
        \includegraphics[width=0.75\linewidth]{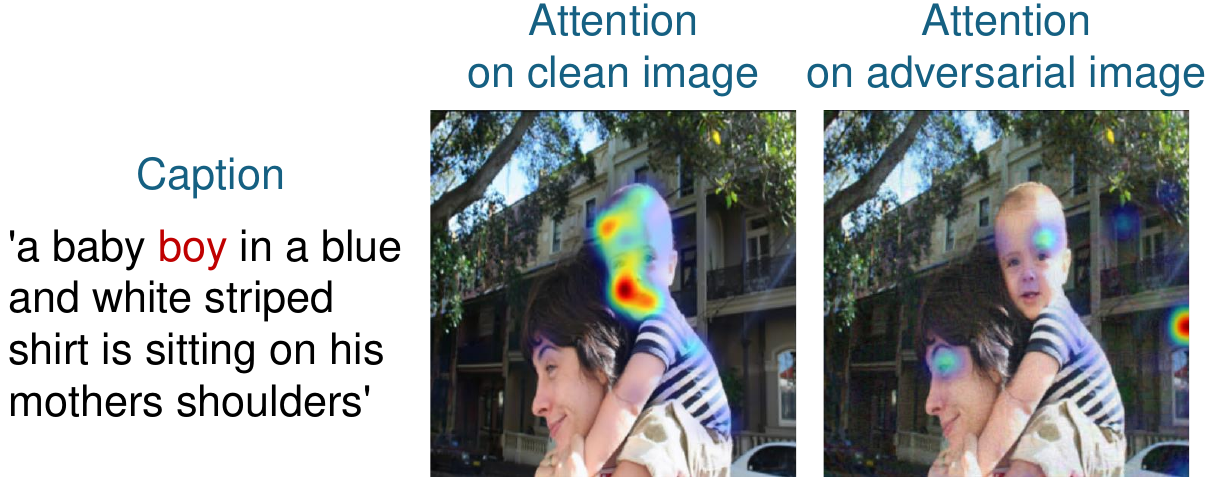}
	\caption{Visualization of visual grounding task.}
	\label{fig:visual_grounding}
\vspace{-10pt}
\end{figure}
For a detailed understanding of the efficiency of our DO-UAP method, we list the detailed time consumption across six VLP models on the Flickr30K and MSCOCO datasets. As shown in Table~\ref{tab:time_comparasion}, the time units for the data are in hours. Across six models, all the methods perform faster on CLIP$_{\text{ViT}}$ and CLIP$_{\text{CNN}}$, while being slower on the other four models. Compared to CPGC-40 (40 epochs), our DO-UAP method is significantly faster overall. Even with the same number of epochs as CPGC-2 (2 epochs), our method demonstrates a slight speed advantage. 

\subsection{Visualization}
We visualize the UAPs generated by our DO-UAP method and their effect on the ITR, IC, and VG tasks as an example in Figure~\ref{fig:ITR_visualization}, Figure~\ref{fig:image_caption}, and Figure~\ref{fig:visual_grounding}. In Figure~\ref{fig:ITR_visualization}(a), given the adversarial query image, we present the ground-truth text alongside the retrieved top-5 texts. It is evident that the retrieved texts differ significantly from the semantics of the query image. Also in Figure~\ref{fig:ITR_visualization}(b), for a given adversarial query text, the top-5 retrieved images are all inconsistent with the query. In Figure~\ref{fig:image_caption}, the VLP model (BLIP) generates a text caption with a meaning that deviates from the actual content of the adversarial image. In Figure~\ref{fig:visual_grounding}, when provided with a text caption and key word, the model is able to correctly identify and locate the ``boy'' in the clean image but fails to do so in the adversarial image.

\subsection{Ablation Study}
\subsubsection{Multimodal loss design}
\begin{table}[tbp]
\centering
\caption{Ablation on weight of unimodal loss.}
\begin{tabular}{c|cccc}
\toprule 
$\alpha$ & 0 & 0.1 & 1 & 10\tabularnewline
\midrule 
TR & 87.68 & 89.04 & 91.75 & 91.63\tabularnewline
IR & 91.90 & 93.94 & 93.81 & 92.28\tabularnewline
\midrule 
Average & 89.79 & 91.49 & \textbf{92.78} & 91.95\tabularnewline
\bottomrule 
\end{tabular}
\label{tab:alpha}
\end{table}
\begin{table}[tbp]
\centering
\caption{Ablation on brightness augmentation.}
\resizebox{\linewidth}{!}{
\begin{tabular}{c|ccccc}
\toprule 
Brightness & N/A & {[}0,0.05{]} & {[}0,0.10{]} & {[}0,0.15{]} & {[}0,0.20{]}\tabularnewline
\midrule 
TR & 91.75 & \textbf{93.35} & 90.27 & 89.29 & 88.67\tabularnewline
IR & 93.81 & \textbf{95.37} & 92.89 & 92.25 & 93.24\tabularnewline
\midrule 
Average & 92.78 & \textbf{94.36} & 91.58 & 90.77 & 90.96\tabularnewline
\midrule 
Image-text similarity & 0.4562 & \textbf{0.4608} & 0.4525 & 0.4550 & 0.4444\tabularnewline
\bottomrule 
\end{tabular}
}
\label{tab:brightness}
\end{table}
\begin{table}[tbp]
\centering
\caption{Ablation of perturbation magnitude per iteration.}
\begin{tabular}{c|ccccc}
\toprule 
$\beta$& 0.1 & 0.3 & 0.5 & 0.7 & 0.9\tabularnewline
\midrule 
TR & 93.35 & 91.38 & 87.68 & 80.79 & 68.84\tabularnewline
IR & 95.37 & 94.80 & 94.29 & 82.07 & 76.94\tabularnewline
\midrule 
Average & \textbf{94.36} & 93.09 & 90.98 & 81.43 & 72.89\tabularnewline
\bottomrule 
\end{tabular}
\label{tab:attack_strength_per_iteration}
\end{table}
\begin{table}[tbp]
\centering
\caption{Ablation study of attack strength.}
\resizebox{0.8\linewidth}{!}{
\begin{tabular}{c|cccc}
\toprule
\multirow{1}{*}{Attack strength} & 4/255 & 8/255 & 12/255 & 16/255\tabularnewline
\midrule
TR & 56.53 & 81.40 & 93.35 & 96.92\tabularnewline
IR & 65.36 & 90.33 & 95.37 & 97.38\tabularnewline
\bottomrule
\end{tabular}
}
\label{tab:attack_strength}
\end{table}
\begin{table*}[tbp]
\centering
\caption{Performance on cross data.}
\resizebox{\linewidth}{!}{
\begin{tabular}{c|c|cccccccccccc|cc}
\toprule
\multirow{2}{*}{Train Dataset} & \multirow{2}{*}{Test Dataset} & \multicolumn{2}{c}{CLIP$_{\text{ViT}}$} & \multicolumn{2}{c}{CLIP$_{\text{CNN}}$} & \multicolumn{2}{c}{BLIP} & \multicolumn{2}{c}{TCL} & \multicolumn{2}{c}{X-VLM} & \multicolumn{2}{c|}{ALBEF} & \multicolumn{2}{c}{Average}\tabularnewline
\cline{3-16}
 &  & TR & IR & TR & IR & TR & IR & TR & IR & TR & IR & TR & IR & TR & IR\tabularnewline
\midrule
Flickr30K & MSCOCO & 96.68 & 96.65 & 97.67 & 97.04 & 93.63 & 92.98 & 97.09 & 94.61 & 97.83 & 94.54 & 97.86 & 96.32 & 96.79 & 95.36\tabularnewline
\midrule
MSCOCO & Flickr30K & 92.49 & 95.12 & 92.62 & 95.34 & 66.77 & 72.92 & 95.55 & 91.57 & 82.42 & 83.01 & 87.36 & 91.34 & 86.20 & 88.22\tabularnewline
\bottomrule
\end{tabular}
}
\label{tab:cross_data}
\end{table*}
\begin{table}[tbp]
\centering
\caption{Performance on image captioning.}
\resizebox{\linewidth}{!}{
\begin{tabular}{c|ccccc}
\toprule
 & B@4 $\downarrow$ & METEOR $\downarrow$ & ROUBE\_L $\downarrow$ & CIDEr $\downarrow$ & SPICE $\downarrow$\tabularnewline
\midrule
BLIP (No attack) & 39.7 & 31.0 & 60.0 & 133.3 & 23.8\tabularnewline
BLIP (Attack) & 24.3 & 23.3 & 48.0 & 82.4 & 16.3\tabularnewline
\bottomrule
\end{tabular}
}
\label{tab:IC}
\end{table}

\begin{table}[tbp]
\centering
\caption{Performance on visual grounding.}
\begin{tabular}{c|c|ccc}
\toprule
Source & Target & Val & TestA & TestB \tabularnewline
\midrule
ALBEF (No attack) & \multirow{4}{*}{ALBEF} & 58.4  & 65.9  & 46.2 \tabularnewline
ALBEF (Attack) &  & 40.7  & 45.8  & 34.1 \tabularnewline
TCL (Attack) &  & 53.3  & 59.2  & 43.1 \tabularnewline
X-VLM (Attack) &  & 56.7  & 62.7  & 44.4 \tabularnewline
\midrule
TCL (No attack) & \multirow{4}{*}{TCL} & 59.6  & 66.8  & 48.1 \tabularnewline
ALBEF (Attack) &  & 56.7  & 63.5  & 46.1 \tabularnewline
TCL (Attack) &  & 49.6  & 54.7  & 40.4 \tabularnewline
X-VLM (Attack) &  & 58.2  & 64.8  & 46.8 \tabularnewline
\midrule
X-VLM (No attack) & \multirow{4}{*}{X-VLM} & 70.8  & 78.2  & 62.5 \tabularnewline
ALBEF (Attack)&  & 65.2  & 69.7  & 58.0 \tabularnewline
TCL (Attack)&  & 65.1  & 70.2  & 57.9 \tabularnewline
X-VLM (Attack)&  & 65.1  & 69.7  & 57.3 \tabularnewline
\bottomrule
\end{tabular}
\label{tab:VG}
\end{table}
For the parameter $\alpha$ in Eq.~\eqref{al:L} which balances the cross-modal and unimodal losses, we conducted an ablation study. The results presented do not include data augmentation, to identify the influence of the unimodal loss.

As shown in Table~\ref{tab:alpha}, the values of $\alpha$ are in the first row. We report the attack success rates for TR (second row) and IR (third row), followed by the average (fourth row). It is evident that when $\alpha$ is set to zero (\ie, without the unimodal loss), the attack performance of UAP is the lowest, underscoring the importance of the unimodal loss. Furthermore, we observe that when $\alpha$ is set to one, the average attack performance is maximized. Therefore, we select $\alpha = 1$ in the final version of our method.

\subsubsection{Data augmentation}
Based on the multimodal objective Eq.~\eqref{con:goal_multi_modal}, we conducted an ablation study to evaluate the impact of brightness augmentation at varying levels of severity. The results highlight the specific influence of this augmentation on attack performance.

As shown in Table~\ref{tab:brightness}, the first row lists the brightness values, representing the range of brightness enhancement applied. We report the attack success rates for TR (second row) and IR (third row), followed by the average (fourth row). It is evident that when the brightness range is [0, 0.05], the attack performance of UAP improves for both TR and IR, with an increase of approximately 2\%. However, excessive augmentation severity does not enhance attack performance. Regarding image-text similarity, only brightness within the range [0, 0.05] leads to an improvement, which aligns with the observed boost in attack performance. This indicates that brightness augmentation, when it improves image-text similarity, also enhances attack effectiveness. Based on the ablation, we select [0, 0.05] brightness in the final version of our method.

\subsubsection{Attack strength per iteration}

Based on the DO-UAP method, we conducted an ablation study to evaluate the impact of perturbation magnitude per iteration ($\beta$). The results reflect the significant influence of $\beta$ on attack performance.

As shown in Table~\ref{tab:attack_strength_per_iteration}, the first row lists the magnitude values, representing the magnitude value in each iteration accounting for how much of the total strength ($\epsilon_v,\epsilon_t$). We report the attack success rates for TR (second row) and IR (third row), followed by the average (fourth row). It is clear that when $\beta$ is set to 0.1, the attack performance of UAP is optimal. Higher values of $\beta$ lead to a decline in performance. Therefore, we select $\beta = 0.1$ in the final version of our method.

\subsubsection{Attack strength}
The magnitude of image perturbations serves as a critical metric for evaluating adversarial attacks. As demonstrated in Table~\ref{tab:attack_strength}, we assessed the attack performance across different perturbation levels, varying from 4/255 to 16/255.

Overall, the attack performance improves as the perturbation magnitude increases. At a perturbation magnitude of 4/255, the performance is notably constrained. This is because, with a limited budget of 4/255, a single common perturbation lacks sufficient capacity to generalize across diverse data samples. Furthermore, from 12/255 to 16/255, the rate of improvement begins to plateau as the perturbation budget grows. 
%

\subsection{Evaluation on More Scenarios}
To further demonstrate the effectiveness of the proposed DO-UAP algorithm, we evaluate its performance in various scenarios, including additional downstream vision-and-language (V+L) tasks, cross-domain transfer, and attacks on state-of-the-art large vision-language models (LVLMs).

\subsubsection{IC}
Image captioning aims to encode an input image into embeddings, which are then used by an image-conditioned language model to generate text descriptions relevant to the image's content. In this work, we utilize BLIP as the victim model, focusing on directly attacking it using the MSCOCO dataset. Following CPGC \cite{fang2024one}, we calculate several standard metrics to assess the quality of the generated captions, including BLEU \cite{papineni2002bleu}, METEOR \cite{banerjee2005meteor}, ROUGE \cite{lin2004rouge}, CIDEr \cite{vedantam2015cider}, and SPICE \cite{anderson2016spice}. All these metrics are lower is better for the attack task. Table~\ref{tab:IC} demonstrates that our DO-UAP method exhibits significant attack effects on the image captioning task.

\begin{figure}[tb]
\centering
        \includegraphics[width=\linewidth]{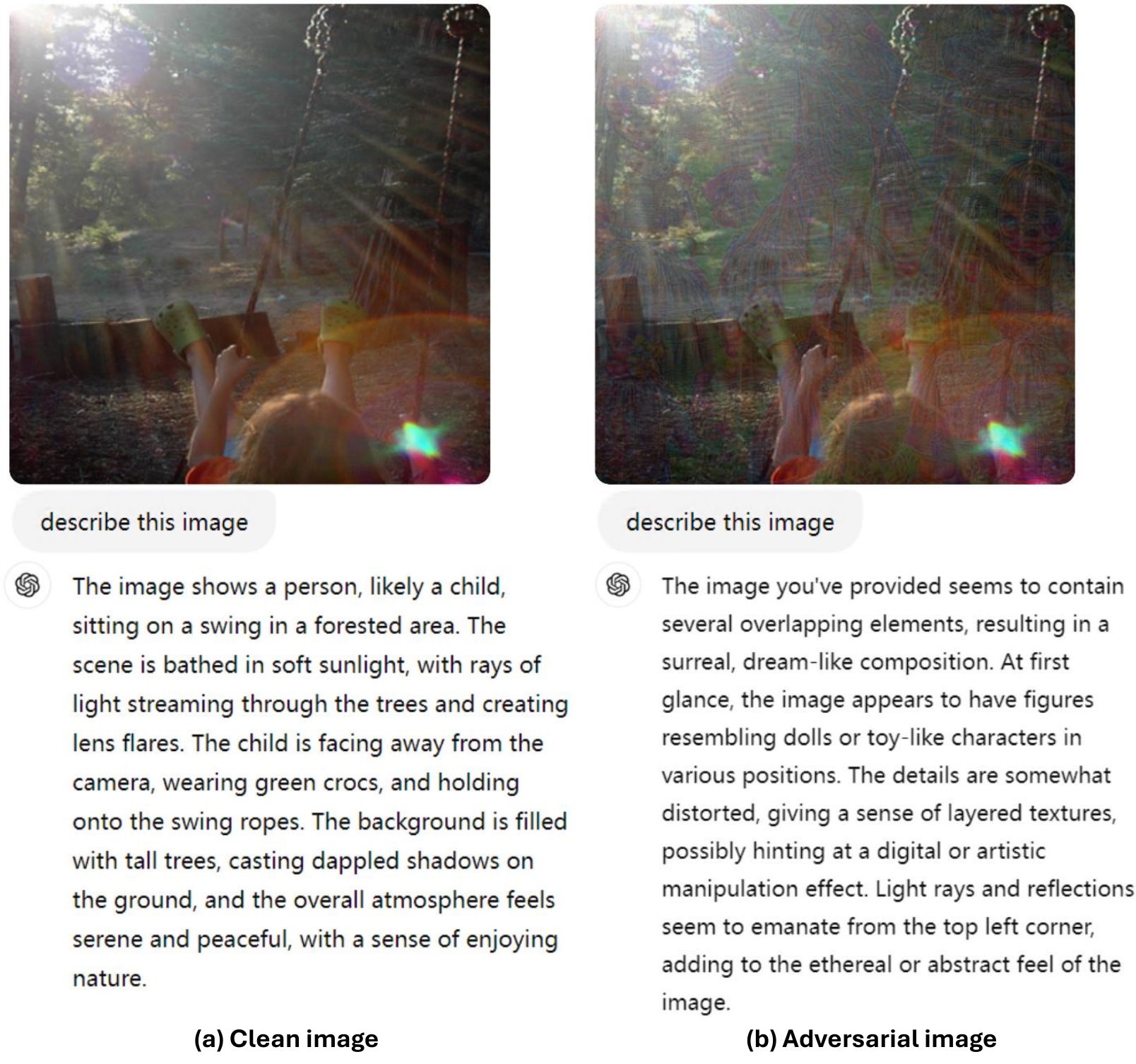}
	\caption{Effect on LVLM ChatGPT 4o.}
	\label{fig:GPT_visual}
\end{figure}

\subsubsection{VG}
Visual grounding is a widely studied vision-and-language (V+L) task that focuses on identifying the correct location in an image based on a given textual description. We perform experiments on the RefCOCO+ dataset, using ALBEF, TCL, and X-VLM as both source and target models. As shown in Table~\ref{tab:VG}, we report the accuracy (lower is better) for each dataset Val, TestA, and TestB. The results demonstrate that our attack effectively reduces localization accuracy, further confirming that the generated UAP effectively disrupts cross-modal interaction and alignment.

\subsubsection{Cross data}
We evaluate the attack performance of the proposed DO-UAP method in a more challenging scenario involving a clear distribution shift between the training and testing data. Specifically, we generate UAP using either the MSCOCO or Flickr30K dataset and assess their transferability to the other. The ASRs are presented in Table~\ref{tab:cross_data}. We can find the UAP generated by our method has strong transferability across different data domains.

\subsubsection{Transfer to LVLMs}
Recently, large general models capable of handling complex vision-language tasks have garnered widespread attention. To further illustrate the potential threat of our method in practical scenarios, \ie, online applications, we aim to evaluate the performance of DO-UAP on mainstream LVLMs ChatGPT 4o. For example, in Figure~\ref{fig:GPT_visual}, the adversarial image with UAP generated by X-VLM can mislead the ChatGPT 4o.

\section{Conclusion}
In this paper, we propose a direct optimization-based UAP approach, termed DO-UAP, which significantly reduces resource consumption while maintaining high attack performance. Specifically, we investigate the necessity of multimodal loss design and introduce an efficient data augmentation strategy. 
%
%
In future work, we aim to enhance the transferability of our algorithm, making the method more practical in a black box setting.

\newpage
{\bibliographystyle{ACM-Reference-Format}
\bibliography{paper_arxiv}}

\newpage
\clearpage
\appendix
\onecolumn
\section{Appendix / supplemental material}\label{sec:appendix}
\subsection{More Results on ITR Task}\label{sec:more_results_ITR}
We present experimental results of the ITR task on R@5 and R@10 in Table~\ref{tab:R5} and Table~\ref{tab:R10}. We report the attack success rate (higher is better) on R@5 and R@10, where R@N denotes the top N most relevant text or image matches based on the query. In summary, our DO-UAP method surpasses CPGC in attack performance with an average improvement of 4.07\% on the TR task and 3.79\% on the IR task across both Flickr30K and MSCOCO datasets. The results are consistent with those in Sec.~\ref{sec:SOTA_comparison}.

\begin{table*}[tbp]
\centering
\caption{Compare with baselines (R@5).}
\resizebox{\linewidth}{!}{
\begin{tabular}{c|c|cccccccccccc|cc|c}
\toprule 
\multirow{2}{*}{Dataset} & \multirow{2}{*}{Methods} & \multicolumn{2}{c}{CLIP$_{\text{ViT}}$} & \multicolumn{2}{c}{CLIP$_{\text{CNN}}$} & \multicolumn{2}{c}{BLIP} & \multicolumn{2}{c}{TCL} & \multicolumn{2}{c}{X-VLM} & \multicolumn{2}{c|}{ALBEF} & \multicolumn{2}{c|}{Average} & \multirow{2}{*}{Time (hours)}\tabularnewline
\cline{3-16}
 &  & TR & IR & TR & IR & TR & IR & TR & IR & TR & IR & TR & IR & TR & IR & \tabularnewline
\midrule 
\multirow{4}{*}{Flickr30K} & GAP & 71.76 & 75.20 & 60.32 & 71.91 & 39.64 & 68.55 & 79.39 & 73.75 & 72.30 & 68.29 & 59.02 & 75.47 & 63.74 & 72.20 & 5.22\tabularnewline
 & CPGC-2 & 81.24 & 85.18 & 61.36 & 72.62 & 53.22 & 65.41 & 85.89 & 78.14 & 67.80 & 69.64 & 78.56 & 73.07 & 71.35 & 74.01 & 6.71\tabularnewline
 & CPGC-40 & 75.85 & 85.76 & 64.74 & 74.37 & 63.48 & 80.53 & \textbf{91.79} & 85.12 & 90.10 & 86.31 & 84.37 & 80.67 & 78.39 & 82.13 & 134.19\tabularnewline
 & DO-UAP (ours) & \textbf{86.42} & \textbf{89.90} & \textbf{83.23} & \textbf{86.45} & \textbf{78.07} & \textbf{86.58} & 89.79 & \textbf{85.30} & \textbf{95.10} & \textbf{90.46} & \textbf{88.68} & \textbf{87.39} & \textbf{86.88} & \textbf{87.68} & 5.83\tabularnewline
\midrule 
\multirow{4}{*}{MSCOCO} & GAP & 95.37 & 93.40 & 89.02 & 90.12 & 64.68 & 68.31 & 93.06 & 88.45 & 91.47 & 83.76 & 75.51 & 79.39 & 84.85 & 83.91 & 5.21\tabularnewline
 & CPGC-2 & 92.78 & 93.76 & 90.32 & 90.42 & 79.66 & 79.70 & 89.15 & 77.57 & 92.79 & 85.09 & 77.17 & 75.40 & 86.98 & 83.66 & 6.51\tabularnewline
 & CPGC-40 & \textbf{96.76} & \textbf{96.57} & 91.95 & 89.34 & \textbf{92.46} & \textbf{93.40} & 95.24 & 91.04 & \textbf{96.95} & \textbf{91.93} & \textbf{93.78} & 91.98 & \textbf{94.52} & 92.38 & 130.13\tabularnewline
 & DO-UAP (ours) & 95.45 & 95.54 & \textbf{93.94} & \textbf{96.35} & 80.77 & 81.21 & \textbf{97.39} & \textbf{93.89} & 96.68 & 93.13 & 93.30 & \textbf{95.36} & 92.92 & \textbf{92.58} & 5.63\tabularnewline
\bottomrule 
\end{tabular}
}
\label{tab:R5}
\end{table*}

\begin{table*}[tbp]
\centering
\caption{Compare with baselines (R@10).}
\resizebox{\linewidth}{!}{
\begin{tabular}{c|c|cccccccccccc|cc|c}
\toprule 
\multirow{2}{*}{Dataset} & \multirow{2}{*}{Methods} & \multicolumn{2}{c}{CLIP$_{\text{ViT}}$} & \multicolumn{2}{c}{CLIP$_{\text{CNN}}$} & \multicolumn{2}{c}{BLIP} & \multicolumn{2}{c}{TCL} & \multicolumn{2}{c}{X-VLM} & \multicolumn{2}{c|}{ALBEF} & \multicolumn{2}{c|}{Average} & \multirow{2}{*}{Time (hours)}\tabularnewline
\cline{3-16}
 &  & TR & IR & TR & IR & TR & IR & TR & IR & TR & IR & TR & IR & TR & IR & \tabularnewline
\midrule 
\multirow{4}{*}{Flickr30K} & GAP & 67.30 & 70.42 & 56.56 & 63.02 & 37.61 & 67.59 & 77.10 & 70.32 & 67.80 & 63.60 & 53.90 & 72.74 & 60.05 & 67.95 & 5.22\tabularnewline
 & CPGC-2 & 77.30 & 79.79 & 52.66 & 64.72 & 46.54 & 58.99 & 83.10 & 72.88 & 61.30 & 64.23 & 72.20 & 67.08 & 65.52 & 67.95 & 6.71\tabularnewline
 & CPGC-40 & 67.88 & 79.55 & 54.60 & 66.58 & 58.88 & 78.29 & \textbf{89.60} & \textbf{81.80} & 87.30 & 83.38 & 81.50 & 76.00 & 73.29 & 77.60 & 134.19\tabularnewline
 & DO-UAP (ours) & \textbf{81.66} & \textbf{86.26} & \textbf{76.79} & \textbf{82.66} & \textbf{72.12} & \textbf{86.58} & 87.50 & 81.05 & \textbf{92.80} & \textbf{88.30} & \textbf{85.60} & \textbf{83.60} & \textbf{82.75} & \textbf{84.74} & 5.83\tabularnewline
\midrule 
\multirow{4}{*}{MSCOCO} & GAP & 93.91 & 91.82 & 87.93 & 86.98 & 60.09 & 66.84 & 91.17 & 85.99 & 89.00 & 80.67 & 71.10 & 76.95 & 82.20 & 81.54 & 5.21\tabularnewline
 & CPGC-2 & 89.89 & 92.15 & 86.94 & 87.00 & 75.47 & 76.25 & 86.16 & 72.42 & 90.30 & 80.75 & 71.49 & 71.29 & 83.38 & 79.98 & 6.51\tabularnewline
 & CPGC-40 & \textbf{95.81} & \textbf{95.55} & 89.42 & 85.81 & \textbf{90.73} & \textbf{92.46} & 94.31 & 89.33 & 95.23 & 89.27 & \textbf{91.97} & 90.21 & \textbf{92.91} & 90.44 & 130.13\tabularnewline
 & DO-UAP (ours) & 94.13 & 93.98 & \textbf{91.99} & \textbf{95.29} & 76.57 & 78.57 & \textbf{96.59} & \textbf{92.21} & \textbf{95.68} & \textbf{91.70} & 90.78 & \textbf{94.30} & 90.96 & \textbf{91.01} & 5.63\tabularnewline
\bottomrule 
\end{tabular}
}
\label{tab:R10}
\end{table*}